\documentclass[letterpaper]{article} 
\usepackage{aaai2027}  
\usepackage[hyphens]{url}  
\usepackage{graphicx} 
\usepackage{multirow}
\usepackage{amsmath} 
\usepackage{adjustbox}
\usepackage{amsmath}
\usepackage{amssymb}
\usepackage{siunitx}
\usepackage{placeins}
\usepackage{natbib}  
\usepackage{caption} 
\usepackage{algorithm}
\usepackage{algorithmic}

\newcommand{\ms}[2]{\ensuremath{#1{\scriptstyle\pm#2}}}

\newcommand{\best}[2]{\ensuremath{\boldsymbol{#1{\scriptstyle\pm#2}}}}
\newcommand{\second}[2]{\ensuremath{\underline{#1{\scriptstyle\pm#2}}}}
\usepackage{newfloat}
\usepackage{listings}
\DeclareCaptionStyle{ruled}{labelfont=normalfont,labelsep=colon,strut=off} 
\floatstyle{ruled}
\newfloat{listing}{tb}{lst}{}
\floatname{listing}{Listing}

\usepackage{booktabs}

\nocopyright 

\title{Breaking Diversity Collapse in Spiking Pseudo-Ensembles for Efficient OOD Detection in Remote Sensing}
\author{
    Srinivas Anumasa, Rushi Shah\textsuperscript{\rm 1}, Qiran Zou\textsuperscript{\rm 1}, Dianbo Liu\textsuperscript{\rm 1}
}
\affiliations{
    \textsuperscript{\rm 1}National University of Singapore\\

}

\begin{document}

\maketitle

\begin{abstract}
Spiking Neural Networks (SNNs) are attractive for resource-constrained
remote-sensing systems, but reliable out-of-distribution (OOD) detection
remains challenging. Deep ensembles provide strong predictive
uncertainty, yet require multiple complete models and backbone
evaluations. We propose an efficient spiking pseudo-ensemble that
attaches multiple lightweight classification heads to a frozen SNN
backbone. Naively training these heads with cross-entropy can lead to
\emph{diversity collapse}, where independently parameterized heads
may produce correlated predictions. To address this, we introduce an agree--disagree objective that
preserves correct predictions on clean in-distribution samples while
encouraging diversity on structured, uncertainty-inducing
 transformations of the same inputs. This provides a diversity-promoting training signal without requiring
external OOD data.
Experiments with Spikformer and ResNet19-SNN on EuroSAT demonstrate
consistent improvements over conventionally trained pseudo-ensembles.
Using three backbones with five heads each matches or improves upon a
five-model deep ensemble on UCM and AID, while requiring approximately
38\% fewer parameters and 40\% fewer backbone evaluations. These
results show that explicit diversity promotion can recover useful
ensemble-style uncertainty at substantially lower deployment cost.
\end{abstract}


\section{Introduction}

\begin{figure}
  \centering
  \includegraphics[width=0.45\textwidth]{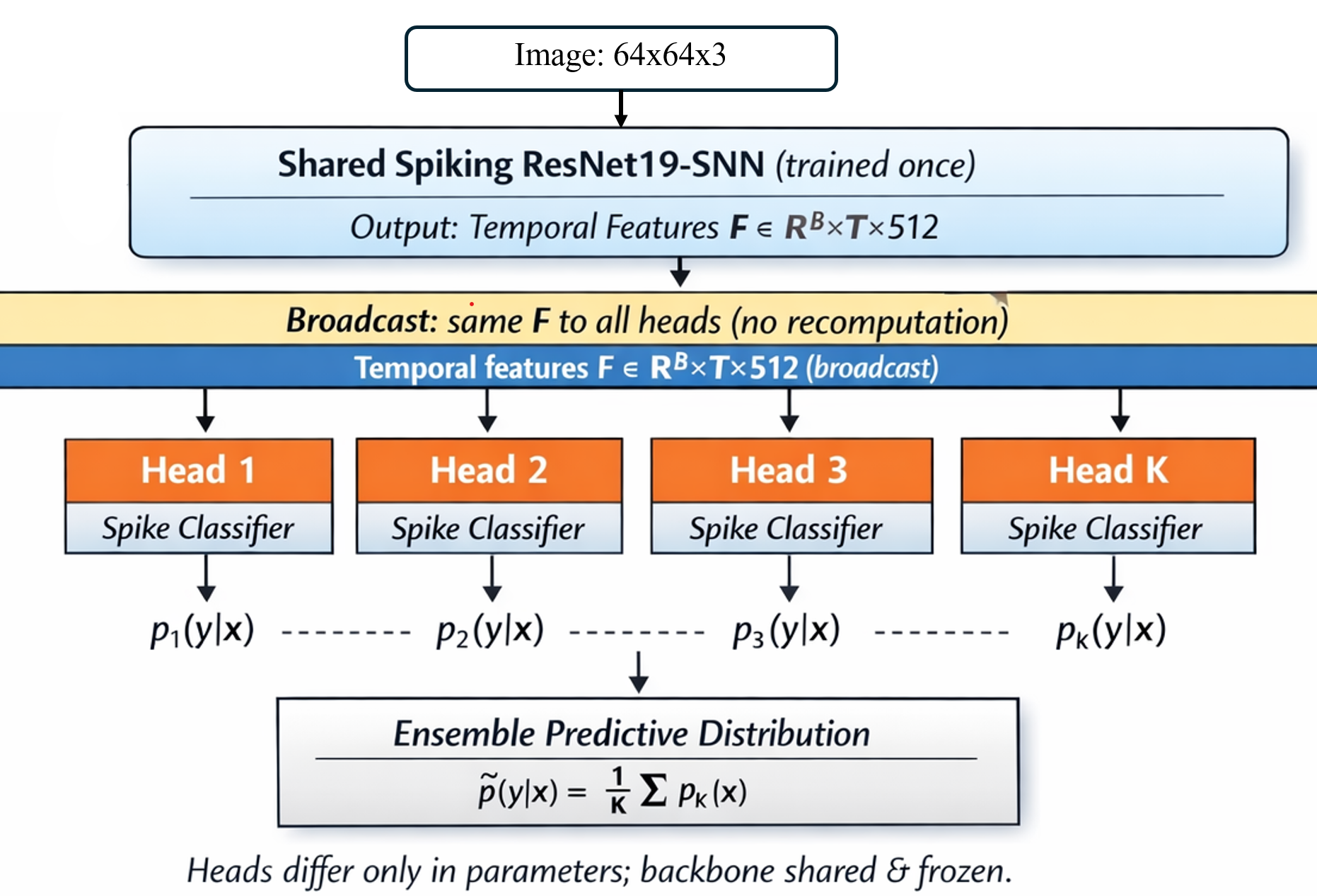}
  \caption{\textbf{Multihead SNN OOD detection.} A shared spiking ResNet19 backbone produces temporal features $F\in\mathbb{R}^{B\times T\times 512}$ which are broadcast to $K_h$ spike-consistent heads. Head predictions are aggregated to form the ensemble predictive distribution, from which MSP, predictive entropy, and mutual information (MI) are computed.}
  \label{fig:multihead_wrap}
\end{figure}
\label{sec:introduction}

Spiking Neural Networks (SNNs) process information through temporally
distributed spike events and can exploit sparse, event-driven
computation. When deployed on compatible neuromorphic hardware, this
computational model can support energy-efficient inference, making SNNs
promising for resource-constrained remote-sensing platforms such as
satellites, unmanned aerial vehicles, and edge sensors
\cite{maass1997networks,davies2018loihi}. Recent advances in surrogate
gradient training, residual SNNs, and spiking self-attention have also
enabled directly trained SNNs to achieve competitive image-classification
performance \cite{neftci2019surrogate,fang2021sew,zhou2022spikformer}.

Reliable deployment, however, requires more than high classification
accuracy. Remote-sensing images can differ substantially from the
training data because of changes in geographic region, season,
illumination, atmospheric conditions, spatial resolution, sensing
platform, or land-cover content. A classifier may nevertheless produce
a highly confident prediction for such unfamiliar inputs. Detecting
out-of-distribution (OOD) samples is therefore essential for deciding
when a prediction should be accepted, rejected, or deferred
\cite{gawlikowski2022advanced,li2024evaluation}.

While uncertainty estimation and OOD detection have been studied
extensively for conventional neural networks, they remain comparatively
underexplored for deep SNNs. Existing SNN research has mainly focused on
classification accuracy, spike efficiency, latency, and training
methods, with only limited work addressing OOD detection through
spike-based representations \cite{martinez2023novel}. Many uncertainty
methods developed for artificial neural networks can, in principle, be
adapted to SNNs. In particular, Monte Carlo dropout and deep ensembles
obtain predictive uncertainty from multiple model realizations
\cite{gal2016dropout,lakshminarayanan2017ensembles}. Deep ensembles
are especially effective because independently trained networks can
represent different predictive functions. Their parameter storage and
inference cost, however, grow linearly with the number of members,
requiring a complete SNN evaluation for every prediction. This repeated
computation can undermine the efficiency that motivates SNN deployment.

Efficient ensemble approximations reduce this cost by sharing parameters
or computations across predictors
\cite{wen2020batchensemble,havasi2020training,
durasov2021masksembles,laurent2023packed}. A particularly simple
alternative is to attach several lightweight classification heads to a
shared pretrained backbone as shown in Figure \ref{fig:multihead_wrap}. Such a pseudo-ensemble requires only one
backbone evaluation to obtain multiple predictions. However,
independently initializing additional heads does not guarantee
functionally diverse predictors. When the heads receive the same
features, labels, and optimization objective, they can converge to
highly correlated predictions, as observed for naive shared-backbone
multihead architectures \cite{havasi2020training}. We refer to this
failure mode as \emph{diversity collapse}.

We address this limitation with an agree--disagree spiking
pseudo-ensemble. A pretrained SNN backbone is frozen and equipped with
multiple lightweight spiking classification heads. The heads are trained
to preserve correct predictions on clean in-distribution samples while
expressing different predictive behaviours on structured,
uncertainty-inducing transformations of the same images. To select these
inputs, we conduct a diagnostic study using independently trained SNNs
and compare several image corruptions. Medium and strong box blur induce
high predictive entropy and inter-model disagreement while retaining
coarse scene structure. We therefore use blur-corrupted inputs to
promote diversity without requiring real OOD samples, an external
outlier dataset, or a pretrained teacher ensemble.

We evaluate the proposed agree--disagree pseudo-ensemble using
Transformer-based Spikformer and convolutional ResNet19-SNN backbones,
with EuroSAT as the in-distribution dataset and UCM, AID, and two
Sentinel-2 Global LULC subsets as distribution shifts. The proposed
training objective consistently improves over cross-entropy-trained
pseudo-ensembles. In particular, a configuration with three backbones and
five heads per backbone matches or exceeds a five-model deep ensemble
on UCM and AID while using approximately $38\%$ fewer parameters and
three-fifths fewer backbone evaluations.

Our main contributions are:

\begin{itemize}
    \item We introduce an efficient spiking pseudo-ensemble that
    generates multiple predictive hypotheses using a frozen SNN backbone
    and lightweight spiking classification heads.

    \item We identify diversity collapse in shared-backbone
    pseudo-ensembles and propose an agree--disagree objective that
    promotes functional diversity using structured blur-corrupted inputs
    derived solely from the ID training data.

    \item We evaluate the method across two distinct SNN architectures
    and multiple remote-sensing distribution shifts, demonstrating an
    improved OOD-performance--deployment-cost trade-off relative to
    conventional pseudo-ensembles and deep ensembles.
\end{itemize}

\section{Background}
\label{sec:background}

This section briefly introduces the spiking-neuron dynamics, input
encoding scheme, and training strategy used by the SNN backbones in
this work.

\subsection{Spiking Neuron Dynamics}

Spiking Neural Networks (SNNs) process information through discrete
spike events distributed over a sequence of simulation steps. Unlike
conventional artificial neurons that communicate using continuous-valued
activations, a spiking neuron maintains an internal membrane potential
and emits a spike when this potential crosses a prescribed threshold
\cite{maass1997networks,gerstner2014neuronal}.

Because remote-sensing images are static, we use constant encoding: the
same normalized RGB image is presented to the network at every
simulation step,
\begin{equation}
\mathbf{x}^{(t)}=\mathbf{x},
\qquad t=1,\ldots,T.
\label{eq:constant_encoding}
\end{equation}
The temporal dynamics of the SNN subsequently transform this repeated
input into a sequence of membrane potentials and spike activations.

We consider the discrete-time leaky integrate-and-fire (LIF) neuron.
For neuron $i$ in layer $l$, its pre-reset membrane potential at time
step $t$ is updated as
\begin{equation}
\widetilde{u}^{(l)}_{i}[t]
=
\beta u^{(l)}_{i}[t-1]
+
\sum_j w^{(l)}_{ij}x^{(l-1)}_{j}[t],
\label{eq:lif_membrane}
\end{equation}
where $u^{(l)}_{i}[t-1]$ is the membrane potential from the previous
time step, $\beta\in[0,1]$ controls membrane leakage,
$w^{(l)}_{ij}$ is the synaptic weight connecting neuron $j$ in layer
$l-1$ to neuron $i$, and $x^{(l-1)}_{j}[t]\in\{0,1\}$ denotes the
presynaptic spike.

The neuron emits a binary spike when its membrane potential exceeds
the firing threshold $u_{\mathrm{th}}$:
\begin{equation}
x^{(l)}_{i}[t]
=
H\!\left(\widetilde{u}^{(l)}_{i}[t]-u_{\mathrm{th}}\right)
=
\begin{cases}
1, & \widetilde{u}^{(l)}_{i}[t] > u_{\mathrm{th}},\\
0, & \text{otherwise},
\end{cases}
\label{eq:lif_spike}
\end{equation}
where $H(\cdot)$ is the Heaviside step function. Following spike
generation, the membrane potential is reset according to
\begin{equation}
u^{(l)}_{i}[t]
=
\widetilde{u}^{(l)}_{i}[t]
\left(1-x^{(l)}_{i}[t]\right).
\label{eq:lif_reset}
\end{equation}

Thus, the membrane potential accumulates temporally distributed input
information, decays according to $\beta$, and is reset when the neuron
fires. The number of simulation steps $T$ controls the temporal duration
of inference and determines an important
accuracy--latency--computation trade-off.

\subsection{Training Deep SNNs}

Training SNNs is challenging because the Heaviside function in
Eq.~\eqref{eq:lif_spike} is non-differentiable. Deep SNNs are commonly
trained either directly using surrogate gradients or obtained through
ANN-to-SNN conversion. In direct training\cite{anumasa2024enhancing,deng2022temporal}, the network is optimized
end-to-end over multiple simulation steps using backpropagation through
time, while the undefined derivative of the spike function is replaced
by the derivative of a smooth surrogate function
\cite{shrestha2018slayer,neftci2019surrogate}. This approach has enabled
deep convolutional and Transformer-based SNN architectures
\cite{fang2021sew,deng2022temporal,zhou2022spikformer}. ANN-to-SNN conversion
instead trains a conventional ANN and subsequently calibrates activation
thresholds, normalization statistics, and reset mechanisms so that SNN
firing rates approximate the original ANN activations
\cite{diehl2015fast,rueckauer2017conversion,deng2021optimal,bojkovic2024data}. Although
conversion can retain high accuracy, it may require more simulation
steps to represent continuous activations accurately.

In this work, both ResNet19-SNN and Spikformer are trained directly
using surrogate-gradient optimization.

\section{Proposed Method}
\label{sec:method}

This section presents the proposed agree--disagree spiking
pseudo-ensemble. We first describe the shared-backbone multihead
architecture and its diversity-collapse problem, and then introduce the
selection of structured disagreement inputs, the agree--disagree
training objective, and the resulting inference-efficiency benefits.
\subsection{Problem Formulation}
\label{sec:problem_formulation}

Let
$\mathcal{D}_{\mathrm{ID}}=\{(x_i,y_i)\}_{i=1}^{N}$
denote an in-distribution remote-sensing training set, where
$x_i\in\mathcal{X}$ is an RGB image and
$y_i\in\{1,\ldots,C\}$ is its scene label. At test time, an input may
originate either from the ID distribution or from an unknown
distribution caused by changes in geography, sensing conditions,
acquisition domain, or semantic content.

Our goal is to obtain multiple predictive hypotheses from a pretrained
SNN while preserving its clean classification performance and avoiding
the cost of a conventional deep ensemble. We assume access only to
$\mathcal{D}_{\mathrm{ID}}$ during training and do not require real OOD
samples or an auxiliary outlier dataset. To promote diversity among the predictors, we construct structured
disagreement inputs directly from the available ID images.

\subsection{Spiking Pseudo-Ensemble Architecture}
\label{sec:pseudo_architecture}

Let $f_{\theta}$ denote a deterministically trained SNN backbone with
parameters $\theta$. Since the remote-sensing images are static, we use
constant encoding, in which the same normalized RGB image is presented
at every simulation step:
\[
\mathbf{x}^{(t)}=\mathbf{x},
\qquad t=1,\ldots,T.
\]

The backbone processes the repeated input over $T$ simulation steps and
produces the time-resolved representation
\begin{equation}
\mathbf{F}_{\theta}(x)
=
\left[
\mathbf{f}^{(1)}(x),
\ldots,
\mathbf{f}^{(T)}(x)
\right]
\in\mathbb{R}^{T\times D},
\label{eq:temporal_features}
\end{equation}
where $D$ is the feature dimension. For Spikformer,
$\mathbf{f}^{(t)}(x)$ is obtained by spatially pooling the output tokens
at simulation step $t$. For ResNet19-SNN, it corresponds to the
time-resolved penultimate representation.

We freeze the backbone parameters and normalization statistics, remove
the original classifier, and attach $K_h$ independently initialized
spiking classification heads
$\{g_{\phi_h}\}_{h=1}^{K_h}$. The same temporal representation is
broadcast to every head. At simulation step $t$, head $h$ computes
\begin{align}
\mathbf{h}^{(t)}_h(x)
&=
\operatorname{LIF}
\left(
\operatorname{BN}_h
\left(
\mathbf{W}_{h,1}\mathbf{f}^{(t)}(x)
+\mathbf{b}_{h,1}
\right)
\right),
\\
\mathbf{z}^{(t)}_h(x)
&=
\mathbf{W}_{h,2}\mathbf{h}^{(t)}_h(x)
+\mathbf{b}_{h,2},
\label{eq:spiking_head}
\end{align}
where $\mathbf{z}^{(t)}_h(x)\in\mathbb{R}^{C}$ denotes the
time-dependent logits.

The temporal logits are averaged before applying the softmax:
\begin{equation}
\overline{\mathbf z}_h(x)
=
\frac{1}{T}\sum_{t=1}^{T}\mathbf z_h^{(t)}(x),
\qquad
\mathbf p_h(x)
=
\operatorname{softmax}\!\left(\overline{\mathbf z}_h(x)\right).
\label{eq:head_probability}
\end{equation}

The predictive distribution of one pseudo-ensemble is obtained by
averaging the probabilities of its $K_h$ heads:
\begin{equation}
\overline{\mathbf{p}}(x)
=
\frac{1}{K_h}
\sum_{h=1}^{K_h}
\mathbf{p}_h(x).
\label{eq:pseudo_ensemble_prediction}
\end{equation}

Unlike a conventional ensemble containing $K_h$ complete networks, the
proposed architecture evaluates the frozen backbone only once and
reuses its temporal representation across $K_h$ lightweight heads.
\subsection{Diversity Collapse under a Frozen Backbone}
\label{sec:diversity_collapse}

A straightforward pseudo-ensemble trains the independently initialized
heads using cross-entropy alone. We refer to this baseline as the
\emph{Cross-Entropy Pseudo-Ensemble} (CEPE). However, introducing additional parameters or independently
initialized heads does not by itself guarantee functionally diverse
predictors. All heads receive the same frozen representation, the same
class labels, and the same optimization objective. Cross-entropy
training can therefore guide them toward similar decision functions,
even when their parameter values are different.

This limitation has also been observed in shared-input ensemble
architectures, where architectural multiplicity alone is insufficient
to produce independent predictive functions
\cite{havasi2020training}. In our setting, parameter sharing is even
stronger because the complete feature extractor is frozen and only the
lightweight classification heads are optimized. Consequently, the
heads may produce highly correlated predictions, providing limited
additional epistemic information as their number increases. We refer
to this failure mode as \emph{diversity collapse}.

The purpose of the proposed agree--disagree objective is therefore not
merely to increase the number of predictors, but to provide an explicit
training signal that encourages the heads to learn different predictive
behaviours away from the clean ID regime.

\begin{table*}[t]
\centering
\caption{
Predictive entropy induced by different corruptions on EuroSAT.
Values are averaged over the evaluation samples; higher values
indicate greater predictive uncertainty. Medium and large box blur
produce the strongest uncertainty response while preserving the
coarse spatial structure of the original scene.
}
\label{tab:corruption_entropy}
\small
\setlength{\tabcolsep}{5.0pt}
\renewcommand{\arraystretch}{1.10}
\begin{tabular}{lcccccccccc}
\toprule
& Clean
& Noise
& Blur-S
& Blur-M
& Blur-L
& Bright.
& Dark
& Contrast
& Rot.\,$90^\circ$
& Cutout
\\
\midrule
Predictive entropy $\uparrow$
& 0.076
& 0.650
& 0.557
& \underline{0.707}
& \textbf{0.752}
& 0.611
& 0.473
& 0.602
& 0.104
& 0.165
\\
\bottomrule
\end{tabular}
\end{table*}
\subsection{Selection of Structured Disagreement Inputs}
\label{sec:blur_selection}

Previous ensemble-diversification methods encourage predictors to
agree on in-distribution data while disagreeing on auxiliary OOD
samples \cite{pagliardini2023dbat,tifrea2022ssnd}. More generally,
Outlier Exposure improves OOD detection by training with an external
dataset of anomalous examples \cite{hendrycks2019oe}. However, suitable
auxiliary outliers may not always be available, particularly in remote
sensing, where future distribution shifts may arise from unknown
sensors, geographic regions, seasons, atmospheric conditions, or
previously unseen land-cover categories.

We therefore seek to construct disagreement inputs directly from the
available ID images. Our goal is not to approximate the unknown
test-time OOD distribution, but to identify structured inputs on which
independently trained predictors naturally express different
hypotheses. Such inputs should satisfy two competing requirements.
First, they should move the predictors away from the high-confidence
clean-data regime and induce meaningful predictive disagreement.
Second, they should preserve sufficient spatial and semantic structure
so that they do not reduce to arbitrary noise.

To identify a suitable transformation, we conduct a diagnostic study
using a conventional deep ensemble of five independently trained
Spikformer models. This study is performed before training the proposed pseudo-ensemble
and examines how different corruption families affect ensemble
uncertainty and inter-model disagreement. Table~\ref{tab:corruption_entropy}
reports the predictive entropy of the ensemble-averaged distribution.
Clean EuroSAT images yield a low mean entropy of $0.076$. Weak
transformations such as $90^{\circ}$ rotation and cutout also leave
the ensemble comparatively confident, with entropy values of $0.104$
and $0.165$, respectively. In contrast, medium and large box blur
increase predictive entropy to $0.707$ and $0.752$, giving the
strongest uncertainty response among the evaluated structured
corruptions.

The same trend is observed in ensemble disagreement. Mutual
information increases from $0.023$ on clean images to $0.263$ and
$0.288$ under medium and large blur, respectively, indicating that
independently trained models produce increasingly different
predictions as blur severity grows. The complete comparison using
MSP, predictive entropy, head variance, and mutual information is
provided in the  supplementary material.

Although additive noise also induces substantial uncertainty, it
introduces unstructured pixel-level perturbations and can severely
distort the natural image statistics. Box blur instead suppresses
fine-grained texture and sharp boundaries while preserving dominant
colour regions and coarse spatial organization. 

These observations suggest that blur naturally exposes regions in
which independently trained models express different predictive
hypotheses. We therefore construct the disagreement set directly from
the ID training images using box blur. For a clean sample $x$, we
generate

\begin{equation}
\tilde{x}=\mathcal{B}_{k}(x),
\end{equation}

where $\mathcal{B}_{k}$ denotes box blurr with kernel size
$k\in\{5,7,9,11\}$. During training, the kernel size is sampled
uniformly for each selected image. The blurred samples are not assumed
to follow the true test-time OOD distribution; rather, they serve as
structured inputs on which the lightweight heads are encouraged to
express different predictive behaviours.

\subsection{Agree--Disagree Training}
\label{sec:agree_disagree}

For each training minibatch, every image is independently selected for
box blur with probability $\rho=0.3$. Let $\mathcal I_{\mathrm{c}}$ denote the
remaining clean ID samples and  $\mathcal I_{\mathrm{b}}$ the blur-corrupted samples.
The two subsets serve complementary purposes: clean samples preserve the
original classification task, whereas blur-corrupted samples provide
structured inputs on which predictive diversity is encouraged.

\paragraph{Agreement on clean ID samples.}
All $K_h$ heads are trained to predict the correct labels for the clean
subset using the average cross-entropy loss,
\[
\mathcal{L}_{\mathrm{CE}}
=
\frac{1}{K_h|\mathcal{I{\mathrm{c}}}|}
\sum_{k=1}^{K_h}
\sum_{i\in\mathcal{I_{\mathrm{c}}}}
\mathrm{CE}\!\left(\mathbf p_k(x_i),y_i\right).
\]
Although this term does not explicitly minimize the distance between
the head predictions, it requires every head to solve the same clean ID
classification task. It therefore preserves agreement where supervised
evidence is available and maintains the classification performance of
the pretrained SNN.

\paragraph{Disagreement on blur-corrupted inputs.}
For each blur-corrupted input $\tilde{x}_i$, we first compute the mean
prediction across the $K_h$ heads,
\[
\bar{\mathbf p}(\tilde{x}_i)
=
\frac{1}{K_h}
\sum_{k=1}^{K_h}
\mathbf p_k(\tilde{x}_i).
\]
We then measure predictive disagreement using the generalized
Jensen--Shannon divergence,
\begin{equation}
\mathcal{D}_{\mathrm{JS}}
=
\frac{1}{K_h|\mathcal{I_{\mathrm{b}}}|}
\sum_{i\in\mathcal{I_{\mathrm{b}}}}
\sum_{k=1}^{K_h}
D_{\mathrm{KL}}
\left(
\mathbf p_k(\tilde{x}_i)
\,\middle\|\,
\bar{\mathbf p}(\tilde{x}_i)
\right).
\label{eq:js_disagreement}
\end{equation}
This quantity is zero when all heads produce identical predictive
distributions and increases as their predictions diverge. Maximizing it
therefore provides an explicit functional-diversity signal that is
absent from ordinary cross-entropy training.

The complete agree--disagree objective is
\begin{equation}
\mathcal{L}_{\mathrm{AD}}
=
\mathcal{L}_{\mathrm{CE}}
-
\lambda_{\mathrm{dis}}\mathcal{D}_{\mathrm{JS}},
\qquad
\lambda_{\mathrm{dis}}=0.3.
\label{eq:agree_disagree_objective}
\end{equation}

During optimization, clean samples contribute only to
$\mathcal{L}_{\mathrm{CE}}$, whereas blur-corrupted samples contribute
only to $\mathcal{D}_{\mathrm{JS}}$. The blurred inputs are therefore
not assigned class labels and are not treated as samples from the true
test-time OOD distributions. Instead, they are used solely to encourage
the heads to express different predictive behaviours away from the clean
training regime.

Only the lightweight classification heads are updated during this
stage; the pretrained SNN backbone and its normalization statistics
remain frozen. We refer to the resulting model as the
\emph{Agree--Disagree Pseudo-Ensemble} (ADPE).

\subsection{Inference and Computational Efficiency}
\label{sec:method_inference}

At inference, each input is evaluated once by every selected backbone,
and its temporal features are passed to the corresponding $K_h$ fixed
heads. With $K_b$ independently trained backbones, the final predictive
distribution is
\begin{equation}
\overline{\mathbf p}_{K_b,K_h}(x)
=
\frac{1}{K_bK_h}
\sum_{b=1}^{K_b}
\sum_{h=1}^{K_h}
\mathbf p_{b,h}(x).
\label{eq:multi_instance_prediction}
\end{equation}
Thus, ADPE produces $K_bK_h$ predictive hypotheses using only $K_b$
backbone evaluations. Its parameter and computation costs are
$P_{\mathrm{ADPE}}=K_b(P_b+K_hP_h)$ and
$\mathcal C_{\mathrm{ADPE}}=K_b(\mathcal C_b+K_h\mathcal C_h)$,
respectively. For both backbones, $P_b\approx12.5$ million and
$P_h\approx0.1$ million; hence, five heads add only about $4\%$
parameter overhead. Since $\mathcal C_h\ll\mathcal C_b$, the dominant
cost scales with $K_b$ rather than $K_bK_h$ work.

\FloatBarrier
\section{Experimental Setup  and Analysis}
\label{sec:experimental_setup}
\begin{table*}[t]
\centering
\caption{
Classification accuracy and MSP-based OOD detection with EuroSAT as
ID. For Mahalanobis, the reported OOD results use its native
feature-distance score rather than MSP.
ADPE denotes our proposed agree--disagree pseudo-ensemble and is
identified by a bold method name.
The pair $(K_b,K_h)$ denotes the number of backbones and heads per
backbone. AUPR treats OOD as the positive class.
All values are percentages.
Higher accuracy, AUROC, and AUPR are better; lower FPR@95 is better.
The best and second-best results within each backbone and metric
column are shown in bold and underlined, respectively.
}
\label{tab:msp_results}

\setlength{\tabcolsep}{3.1pt}
\renewcommand{\arraystretch}{1.07}

\begin{tabular}{@{}lccccccc@{}}
\toprule
\multirow{2}{*}{Method}
& \multirow{2}{*}{Acc.$\uparrow$}
& \multicolumn{3}{c}{UCM}
& \multicolumn{3}{c}{AID}
\\
\cmidrule(lr){3-5}
\cmidrule(lr){6-8}
&
& AUROC$\uparrow$
& AUPR$\uparrow$
& FPR@95$\downarrow$
& AUROC$\uparrow$
& AUPR$\uparrow$
& FPR@95$\downarrow$
\\
\midrule

\multicolumn{8}{c}{\textbf{Spikformer}} \\
\midrule

Maha. $(1,\text{--})$
& \ms{97.22}{0.18}
& \ms{95.09}{0.51}
& \ms{94.20}{0.80}
& \ms{22.97}{2.45}
& \ms{95.16}{0.68}
& \ms{98.59}{0.21}
& \ms{22.09}{1.34}
\\

MC-DO $(1,\text{--})$
& \ms{97.61}{0.11}
& \ms{92.13}{0.48}
& \ms{90.86}{0.59}
& \ms{34.41}{2.59}
& \ms{93.70}{0.99}
& \ms{98.08}{0.29}
& \ms{30.56}{3.78}
\\

LLL $(1,\text{--})$
& \ms{97.47}{0.17}
& \ms{91.53}{1.37}
& \ms{89.48}{1.44}
& \ms{39.37}{4.29}
& \ms{92.53}{1.28}
& \ms{97.53}{0.48}
& \ms{37.33}{6.44}
\\

DE $(1,1)$
& \ms{97.58}{0.10}
& \ms{89.53}{1.66}
& \ms{87.28}{1.58}
& \ms{43.71}{3.40}
& \ms{90.57}{1.97}
& \ms{96.87}{0.67}
& \ms{42.63}{6.44}
\\

DE $(3,1)$ & $98.23\pm0.08$ & $94.05\pm0.53$ & $92.72\pm0.57$ & $27.57\pm2.28$ & $94.70\pm0.35$ & $98.27\pm0.12$ & $27.07\pm2.25$ \\

DE $(5,1)$
& \best{98.44}{0.00}
& \ms{94.84}{0.00}
& \ms{93.69}{0.00}
& \ms{23.71}{0.00}
& \ms{95.41}{0.00}
& \ms{98.50}{0.00}
& \ms{22.93}{0.00}

\\

CEPE $(3,5)$
& \second{98.40}{0.11}
& \ms{95.10}{0.27}
& \ms{94.47}{0.32}
& \ms{22.32}{1.11}
& \ms{95.97}{0.29}
& \ms{98.79}{0.10}
& \ms{19.98}{1.57}
\\
\midrule

\textbf{ADPE $(1,5)$}
& \ms{97.56}{0.21}
& \ms{94.82}{0.60}
& \ms{94.13}{0.75}
& \ms{23.41}{2.20}
& \ms{95.33}{1.05}
& \ms{98.60}{0.36}
& \ms{22.52}{5.46}
\\

\textbf{ADPE $(2,5)$}
& \ms{98.16}{0.16}
& \second{96.47}{0.46}
& \second{95.84}{0.51}
& \second{16.87}{1.77}
& \second{96.73}{0.46}
& \second{99.03}{0.17}
& \second{15.24}{2.72}
\\

\textbf{ADPE $(3,5)$}
& \ms{98.30}{0.12}
& \best{97.12}{0.30}
& \best{96.61}{0.29}
& \best{14.75}{0.82}
& \best{97.29}{0.24}
& \best{99.22}{0.09}
& \best{13.07}{1.06}
\\

\midrule
\multicolumn{8}{c}{\textbf{ResNet19-SNN}} \\
\midrule

Maha. $(1,\text{--})$
& \ms{97.86}{0.27}
& \ms{83.43}{3.01}
& \ms{78.33}{3.62}
& \ms{65.47}{5.32}
& \ms{77.67}{4.42}
& \ms{90.99}{2.34}
& \ms{79.06}{8.25}
\\

MC-DO $(1,\text{--})$
& \ms{98.04}{0.23}
& \ms{88.90}{1.17}
& \ms{87.14}{1.39}
& \ms{42.51}{5.30}
& \ms{90.34}{0.84}
& \ms{96.93}{0.31}
& \ms{40.09}{4.55}
\\

LLL $(1,\text{--})$
& \ms{98.10}{0.23}
& \ms{89.45}{0.43}
& \ms{87.29}{0.76}
& \ms{43.13}{2.95}
& \ms{90.40}{0.50}
& \ms{96.91}{0.18}
& \ms{41.08}{3.74}
\\

DE $(1,1)$
& \ms{98.08}{0.10}
& \ms{88.96}{0.95}
& \ms{86.45}{1.23}
& \ms{44.32}{3.20}
& \ms{90.00}{0.53}
& \ms{96.69}{0.20}
& \ms{42.50}{2.87}
\\
DE $(3,1)$ & $98.49\pm0.07$ & $91.27\pm0.28$ & $89.65\pm0.40$ & $36.33\pm1.35$ & $92.25\pm0.13$ & $97.53\pm0.06$ & $34.77\pm1.52$ \\
DE $(5,1)$
& \best{98.56}{0.00}
& \second{91.84}{0.00}
& \ms{90.38}{0.00}
& \second{33.62}{0.00}
& \second{92.79}{0.00}
& \second{97.72}{0.00}
& \second{31.82}{0.00}
\\

CEPE $(3,5)$
& \second{98.50}{0.13}
& \ms{91.30}{0.19}
& \ms{90.00}{0.31}
& \ms{34.33}{1.47}
& \ms{92.59}{0.14}
& \ms{97.70}{0.06}
& \ms{31.91}{1.57}
\\
\midrule

\textbf{ADPE $(1,5)$}
& \ms{98.08}{0.10}
& \ms{89.90}{0.70}
& \ms{88.58}{1.06}
& \ms{39.19}{3.13}
& \ms{91.02}{0.45}
& \ms{97.17}{0.17}
& \ms{38.73}{1.87}
\\

\textbf{ADPE $(2,5)$}
& \ms{98.40}{0.16}
& \ms{91.51}{0.30}
& \second{90.44}{0.51}
& \ms{34.33}{1.98}
& \ms{92.55}{0.26}
& \ms{97.69}{0.11}
& \ms{33.42}{1.69}
\\

\textbf{ADPE $(3,5)$}
& \second{98.50}{0.14}
& \best{92.10}{0.17}
& \best{91.17}{0.34}
& \best{32.03}{1.05}
& \best{93.12}{0.17}
& \best{97.89}{0.07}
& \best{31.07}{1.28}
\\

\bottomrule
\end{tabular}
\end{table*}

\paragraph{Datasets.}
We use EuroSAT \cite{helber2019eurosat} as the in-distribution dataset
and evaluate OOD detection on UCM \cite{yang2010bag},
AID \cite{xia2017aid}, and two subsets of Sentinel-2 Global LULC
\cite{benhammou2022sentinel}.
Global-Near contains classes with semantic overlap with EuroSAT,
whereas Global-Far contains classes without a direct EuroSAT
counterpart. All images are resized to $64\times64$ and normalized
using ImageNet statistics.

\paragraph{Models and training.}
We evaluate Transformer-based Spikformer and convolutional
ResNet19-SNN backbones, both directly trained with $T=2$ simulation
steps. Five independently initialized backbones are trained for each
architecture. After deterministic training, each backbone is frozen
and equipped with $K_h=5$ lightweight spiking heads. CEPE trains these
heads using cross-entropy alone, whereas ADPE additionally applies the
agree--disagree objective in Eq.~\eqref{eq:agree_disagree_objective}.
Complete architectural and optimization details are provided in the
supplementary material.

\paragraph{Baselines and evaluation.}
We compare ADPE against a deterministic SNN, deep ensembles, Monte
Carlo dropout, last-layer Laplace, Mahalanobis distance,  and the
cross-entropy pseudo-ensemble CEPE.A configuration $(K_b,K_h)$ denotes
the number of selected backbones and the number of heads per backbone,
respectively. For each $K_b$, we evaluate all
$\binom{5}{K_b}$ possible subsets of the five trained backbones and
report the mean and standard deviation across these subsets. For $K_b=5, k_h=1$, only one subset exists, so the reported standard deviation is zero. We report EuroSAT classification accuracy and OOD
detection using MSP and mutual information. Performance is measured
using AUROC, AUPR-Out, and FPR@95, where FPR@95 is the fraction of OOD
samples accepted as ID at a threshold retaining $95\%$ of the ID
samples. Full baseline implementations, uncertainty-score definitions,
and reproducibility details are provided in the supplementary material.



\begin{table*}[t]
\centering
\caption{
OOD detection using mutual information (MI) with EuroSAT as ID.
ADPE denotes our proposed agree--disagree pseudo-ensemble and is
identified by a bold method name.
The pair $(K_b,K_h)$ denotes the number of independently trained
backbones and the number of heads per backbone.
Results are mean$\pm$standard deviation and are reported as
percentages.
Higher AUROC is better; lower FPR@95 is better.
The best and second-best results within each backbone and metric
column are shown in bold and underlined, respectively.
}
\label{tab:mi_results}

\setlength{\tabcolsep}{2.4pt}
\renewcommand{\arraystretch}{1.07}

\begin{tabular}{@{}lcccccccc@{}}
\toprule
\multirow{2}{*}{Method}
& \multicolumn{2}{c}{UCM}
& \multicolumn{2}{c}{AID}
& \multicolumn{2}{c}{Global-Near}
& \multicolumn{2}{c}{Global-Far}
\\
\cmidrule(lr){2-3}
\cmidrule(lr){4-5}
\cmidrule(lr){6-7}
\cmidrule(lr){8-9}
& AUROC$\uparrow$ & FPR@95$\downarrow$
& AUROC$\uparrow$ & FPR@95$\downarrow$
& AUROC$\uparrow$ & FPR@95$\downarrow$
& AUROC$\uparrow$ & FPR@95$\downarrow$
\\
\midrule

\multicolumn{8}{c}{\textbf{Spikformer}} \\
\midrule

MC-DO $(1,\text{--})$
& \ms{85.94}{0.90}
& \ms{46.30}{4.47}
& \ms{85.73}{1.30}
& \ms{46.03}{4.92}
& \ms{78.25}{3.24}
& \ms{51.87}{5.70}
& \ms{72.37}{1.60}
& \ms{56.19}{4.40}
\\

LLL $(1,\text{--})$
& \ms{92.98}{1.40}
& \ms{32.03}{3.83}
& \ms{93.94}{1.22}
& \ms{30.38}{4.56}
& \ms{87.38}{2.06}
& \ms{49.96}{8.63}
& \ms{89.85}{2.65}
& \ms{50.02}{6.70}
\\

DE $(5,1)$
& \ms{95.18}{0.00}
& \second{22.43}{0.00}
& \ms{95.64}{0.00}
& \ms{22.53}{0.00}
& \ms{90.53}{0.00}
& \best{35.24}{0.00}
& \second{93.66}{0.00}
& \ms{40.32}{0.00}
\\

CEPE $(3,5)$
& \ms{93.25}{0.31}
& \ms{33.80}{4.04}
& \ms{94.40}{0.29}
& \ms{28.37}{3.05}
& \ms{90.48}{1.20}
& \ms{40.27}{2.45}
& \ms{92.01}{0.49}
& \ms{44.57}{2.06}
\\
\midrule

\textbf{ADPE $(1,5)$}
& \ms{94.67}{0.56}
& \best{21.90}{1.94}
& \ms{94.89}{0.83}
& \ms{21.57}{4.53}
& \ms{88.33}{1.12}
& \ms{47.36}{7.49}
& \ms{90.43}{1.60}
& \ms{39.10}{2.86}
\\

\textbf{ADPE $(2,5)$}
& \second{95.51}{0.55}
& \ms{25.06}{3.55}
& \second{95.99}{0.37}
& \second{20.66}{2.91}
& \second{91.43}{0.79}
& \ms{40.33}{5.30}
& \ms{93.46}{1.12}
& \second{33.01}{4.57}
\\

\textbf{ADPE $(3,5)$}
& \best{95.97}{0.45}
& \ms{25.83}{3.40}
& \best{96.45}{0.23}
& \best{20.54}{2.40}
& \best{92.54}{0.44}
& \second{36.88}{2.88}
& \best{94.47}{0.70}
& \best{30.70}{4.08}
\\

\midrule
\multicolumn{8}{c}{\textbf{ResNet19-SNN}} \\
\midrule

MC-DO $(1,\text{--})$
& \ms{88.24}{1.15}
& \ms{44.50}{4.21}
& \ms{89.59}{0.78}
& \ms{42.23}{3.61}
& \ms{88.62}{0.98}
& \ms{43.48}{10.08}
& \ms{92.67}{1.14}
& \ms{39.84}{4.83}
\\

LLL $(1,\text{--})$
& \ms{89.42}{0.51}
& \ms{43.09}{2.34}
& \ms{90.42}{0.57}
& \ms{40.90}{2.68}
& \ms{88.23}{1.26}
& \ms{42.55}{6.28}
& \ms{92.50}{1.30}
& \ms{38.71}{4.73}
\\

DE $(5,1)$
& \ms{91.84}{0.00}
& \ms{34.10}{0.00}
& \second{92.85}{0.00}
& \best{33.03}{0.00}
& \second{90.44}{0.00}
& \best{31.15}{0.00}
& \second{96.12}{0.00}
& \best{19.80}{0.00}
\\

CEPE $(3,5)$
& \ms{91.34}{0.25}
& \ms{35.66}{1.51}
& \ms{92.55}{0.17}
& \second{33.12}{1.31}
& \ms{89.71}{0.52}
& \second{35.96}{4.39}
& \ms{95.64}{0.52}
& \ms{23.12}{5.09}
\\
\midrule

\textbf{ADPE $(1,5)$}
& \ms{90.84}{0.79}
& \ms{38.11}{3.84}
& \ms{90.44}{0.89}
& \ms{41.34}{4.23}
& \ms{88.82}{1.30}
& \ms{49.88}{6.29}
& \ms{93.37}{0.78}
& \ms{35.98}{5.37}
\\

\textbf{ADPE $(2,5)$}
& \second{92.11}{0.27}
& \second{34.02}{2.32}
& \ms{92.21}{0.32}
& \ms{35.92}{1.73}
& \ms{89.93}{0.82}
& \ms{43.42}{4.18}
& \ms{95.41}{0.59}
& \ms{26.10}{4.99}
\\

\textbf{ADPE $(3,5)$}
& \best{92.68}{0.11}
& \best{31.92}{1.02}
& \best{92.92}{0.20}
& \ms{33.34}{1.00}
& \best{90.58}{0.59}
& \ms{39.48}{3.50}
& \best{96.13}{0.32}
& \second{21.72}{2.20}
\\

\bottomrule
\end{tabular}
\end{table*}

\subsection{Overall OOD Detection Performance}

Table~\ref{tab:msp_results} reports ID classification accuracy and
MSP-based OOD detection on UCM and AID. Across both backbones, ADPE
improves as additional independently trained backbone instances are
combined, while retaining essentially the same EuroSAT classification
accuracy as CEPE and conventional deep ensembles.

For Spikformer, ADPE with three backbones achieves the strongest
results across all reported UCM and AID detection metrics. On UCM,
ADPE $(3,5)$ obtains an AUROC of $97.12\%$ and an FPR@95 of
$14.75\%$, compared with $94.84\%$ and $23.71\%$, respectively, for
the five-model deep ensemble. On AID, it similarly improves AUROC
from $95.41\%$ to $97.29\%$ and reduces FPR@95 from $22.93\%$ to
$13.07\%$. These improvements are obtained while using only three
backbone evaluations instead of five.

The same overall trend is observed for ResNet19-SNN, although the
performance differences are smaller. ADPE $(3,5)$ achieves AUROC
values of $92.10\%$ and $93.12\%$ on UCM and AID, respectively,
slightly exceeding the five-model deep ensemble values of $91.84\%$
and $92.79\%$. It also reduces FPR@95 from $33.62\%$ to $32.03\%$
on UCM and from $31.82\%$ to $31.07\%$ on AID. Thus, the benefit of
the proposed objective is observed for both Transformer-based and
convolutional SNNs, but is more pronounced for Spikformer.

Importantly, these gains do not come at the expense of ID
classification performance. ADPE $(3,5)$ obtains EuroSAT accuracies
of $98.30\%$ and $98.50\%$ for Spikformer and ResNet19-SNN,
respectively. These values are within $0.14$ percentage points of the
corresponding five-model deep ensembles, indicating that promoting
disagreement on blurred inputs does not substantially compromise the
clean classification task.

\subsection{Effect of Agree--Disagree Training}

The comparison between CEPE and ADPE isolates the effect of the
disagreement objective because both methods use the same frozen
backbones, head architectures, optimization settings, and number of
heads. The only difference is the Jensen--Shannon disagreement term
applied to blur-corrupted inputs.

For Spikformer, ADPE $(3,5)$ improves over CEPE $(3,5)$ by
$2.02$ percentage points in UCM AUROC and reduces FPR@95 by
$7.57$ points. On AID, the corresponding improvements are
$1.32$ AUROC points and $6.91$ FPR@95 points. For ResNet19-SNN,
ADPE improves UCM AUROC from $91.30\%$ to $92.10\%$ and reduces
FPR@95 from $34.33\%$ to $32.03\%$. On AID, it improves AUROC from
$92.59\%$ to $93.12\%$ and reduces FPR@95 from $31.91\%$ to
$31.07\%$.

The single-backbone setting provides further evidence that the gains
are not explained only by combining independently trained models.
For Spikformer, ADPE $(1,5)$ improves UCM AUROC from $89.53\%$ for
the deterministic model to $94.82\%$, while reducing FPR@95 from
$43.71\%$ to $23.41\%$. Similar improvements are obtained on AID.
The gain is smaller but still visible for ResNet19-SNN. These results
show that five heads attached to one frozen backbone can provide
useful additional predictive variation when they are explicitly
trained to avoid collapsing to the same behavior.

\subsection{Scaling the Number of Backbones}

Increasing $K_b$ consistently improves MSP-based detection for ADPE.
For Spikformer, UCM AUROC increases from $94.82\%$ at $K_b=1$ to
$96.47\%$ at $K_b=2$ and $97.12\%$ at $K_b=3$. Over the same
configurations, FPR@95 decreases from $23.41\%$ to $16.87\%$ and
then to $14.75\%$. A similar monotonic trend is observed on AID and
for ResNet19-SNN.

This result illustrates the complementary roles of within-backbone
and across-backbone diversity. The multiple heads provide several
predictive hypotheses for each frozen representation, whereas
independently trained backbones introduce additional variation in the
representation itself. Combining both sources produces stronger OOD
detection than increasing either source alone.

Since five heads add only approximately $4\%$ parameter overhead per
backbone, ADPE $(3,5)$ requires approximately $3.12$ full-model
parameter equivalents, compared with five for DE $(5,1)$. It therefore
uses approximately $38\%$ less parameter storage and $40\%$ fewer
backbone evaluations, while matching or improving its
OOD detection performance on UCM and AID. These quantities represent
model-size and computation proxies rather than direct hardware-energy
measurements.

\subsection{Predictive Disagreement Analysis}

Table~\ref{tab:mi_results} evaluates mutual information, which directly
measures disagreement among the predictive members. The comparison
between CEPE and ADPE is particularly informative because a
cross-entropy-trained pseudo-ensemble may achieve reasonable MSP
performance while its heads still produce highly correlated
predictions.

For Spikformer, ADPE $(3,5)$ improves MI-based AUROC over CEPE
$(3,5)$ by $2.72$, $2.05$, $2.06$, and $2.46$ percentage points on
UCM, AID, Global-Near, and Global-Far, respectively. For
ResNet19-SNN, the corresponding improvements are $1.34$, $0.37$,
$0.87$, and $0.49$ points. The improvement across all evaluated
datasets is consistent with the intended effect of the disagreement
objective: the heads retain similar clean accuracy but produce more
informative predictive differences away from the clean training
distribution.

ADPE $(3,5)$ also achieves the highest MI AUROC for nearly every
dataset and backbone configuration. For Spikformer, it exceeds the
five-model deep ensemble on UCM, AID, Global-Near, and Global-Far.
For ResNet19-SNN, it is either slightly better than or comparable to
the five-model ensemble across the four datasets.

The FPR@95 results, however, are less uniformly monotonic. For
example, Spikformer ADPE $(1,5)$ obtains a lower UCM MI-based FPR@95
than ADPE configurations with more backbones, even though the latter
have higher AUROC. Similarly, the five-model ResNet19-SNN ensemble
retains the lowest FPR@95 on some Global-LULC shifts. This indicates
that improved global ranking, as measured by AUROC, does not always
produce the best operating point at a threshold fixed to accept
$95\%$ of ID samples. The location and spread of the uncertainty-score
distributions therefore remain important, particularly for
semantically close distribution shifts.

\subsection{Effect of OOD Shift Type}

Detection is generally more difficult on Global-Near than on
Global-Far. For example, Spikformer ADPE $(3,5)$ obtains an MI AUROC
of $92.54\%$ on Global-Near and $94.47\%$ on Global-Far. A similar
difference is observed for ResNet19-SNN. This behavior is consistent
with the construction of the subsets: Global-Near contains classes
that overlap semantically with EuroSAT and are therefore more likely
to occupy regions close to the ID representation, whereas
Global-Far contains classes without a direct EuroSAT counterpart.

Although ADPE improves MI AUROC on both shifts, its FPR@95 advantage
is less consistent on Global-Near. The proposed method should
therefore be viewed as an efficient mechanism for increasing useful
ensemble disagreement rather than as a uniformly superior detector
under every shift and operating threshold.

\section{Conclusion}

We introduced an efficient spiking pseudo-ensemble for remote-sensing
OOD detection. The proposed agree--disagree objective addresses
diversity collapse by preserving classification performance on clean ID
samples while encouraging lightweight heads to express different
predictions on structured disagreement inputs. Across Spikformer and
ResNet19-SNN, the method improves over conventionally trained
pseudo-ensembles and provides a favorable trade-off relative to deep
ensembles, requiring fewer backbone evaluations and substantially less
parameter replication. The results also indicate that semantically
close distribution shifts and performance at fixed operating thresholds
remain challenging, motivating further study of adaptive disagreement
inputs and uncertainty calibration in SNNs.
\bibliography{aaai2027}



\clearpage
\section*{Supplementary Material Overview}

This document provides additional dataset details, implementation and
training settings, random-seed configuration, computational resources,
formal definitions of the evaluation metrics, and the trainable-parameter
calculation supporting the results reported in the main paper.
\label{sec:supp_experimental_details}

\subsection{Datasets and Preprocessing}
\label{sec:supp_datasets}

All experiments use \textbf{EuroSAT}~\cite{helber2019eurosat} as the
in-distribution (ID) dataset. EuroSAT contains 27,000 Sentinel-2 images
from 10 land-use and land-cover classes. We use only the RGB channels,
resize every image to $64\times64$ pixels, and normalize the channels
using the ImageNet mean and standard deviation. The dataset is
partitioned using a fixed train/validation/test split stored in a single
split file. The same split is used for every architecture, baseline,
and random seed.


For OOD evaluation, we use UCMerced Land Use
(UCM)~\cite{yang2010bag}, AID~\cite{xia2017aid}, and Sentinel-2
Global Land Use/Land Cover~\cite{benhammou2022sentinel}. UCM and
AID are evaluated using their complete datasets. For Sentinel-2 Global
LULC, we construct two subsets according to semantic overlap with the
EuroSAT classes. Global-Near contains 21 classes with semantic overlap
with EuroSAT, whereas Global-Far contains eight classes without an
apparent EuroSAT counterpart. To reduce class-size imbalance, every
selected Global LULC class is capped at 416 images, corresponding to
the size of the smallest selected class.


During backbone training, we apply random horizontal flipping and a
random resized crop with scale range $[0.8,1.0]$. Validation, ID test,
and OOD images are resized and normalized without stochastic
augmentation.

\subsection{SNN Backbones and Classification Heads}
\label{sec:supp_architectures}

We evaluate two directly trained SNN architectures: a Transformer-based
Spikformer and a convolutional ResNet19-SNN. Both use $T=2$ simulation
steps.

\paragraph{Spikformer.}\cite{zhou2022spikformer}
The Spikformer uses six Transformer blocks, embedding dimension 384,
six attention heads, patch size 4, and an MLP expansion ratio of 4.
Spiking activations are implemented using
\texttt{MultiStepLIFNode} from SpikingJelly with membrane time constant
$\tau=2.0$ and \texttt{detach\_reset=True}. We use the sigmoid
surrogate gradient with slope parameter $\alpha=4.0$. Each
pseudo-ensemble head has the form
\[
\mathrm{Linear}(384,256)
\rightarrow \mathrm{BN}
\rightarrow \mathrm{LIF}
\rightarrow \mathrm{Linear}(256,C).
\]

\paragraph{ResNet19-SNN.}\cite{deng2022temporal}
The ResNet19-SNN uses residual-layer configuration $[3,3,2]$ and
channel dimensions
$64\rightarrow128\rightarrow256\rightarrow512$. The network uses
LIFSpike neurons with membrane decay $\tau=0.5$, a piecewise-linear
surrogate gradient with $\gamma=1.0$, and a hard membrane reset to zero
at the beginning of each forward pass. Each pseudo-ensemble head is
\[
\mathrm{tdLayer}(512,256)
\rightarrow \mathrm{LIFSpike}
\rightarrow \mathrm{tdLayer}(256,C).
\]

\subsection{Backbone and Pseudo-Ensemble Training}
\label{sec:supp_training}

Each deterministic backbone is trained for 300 epochs using
cross-entropy loss. After deterministic training, the checkpoint with
the highest validation accuracy is retained. Its backbone parameters
and normalization statistics are then frozen, the original classifier
is removed, and $K_h=5$ independently initialized classification heads
are attached. 

The heads are trained simultaneously for 100 epochs. For CEPE, each
head receives minibatches from its own independently shuffled data
loader and is updated using a separate optimizer. This provides a
stronger cross-entropy baseline by exposing the heads to different
minibatch sequences. Although the heads may receive different samples
at a particular optimization step, they share the same frozen backbone
and therefore operate in the same representation space.

ADPE uses a shared minibatch because the generalized
Jensen--Shannon objective requires predictions from all heads for the
same input. Each ADPE head is nevertheless updated using a separate
optimizer. All head-specific optimizers use identical hyperparameters
and learning-rate schedules.

For ADPE, each training image is independently selected for box blur
with probability $\rho=0.3$. Blur is implemented using average pooling
with kernel size
\[
k\in\{5,7,9,11\},
\]
sampled uniformly for each selected image. Clean images contribute only
to the cross-entropy objective, whereas blur-corrupted images are
excluded from cross-entropy and contribute only to the generalized
Jensen--Shannon disagreement objective, with
$\lambda_{\mathrm{dis}}=0.3$. Therefore, no class label is imposed on a
blur-corrupted image.

For both CEPE and ADPE, each classification head is updated using a
separate optimizer. All head-specific optimizers use the same
optimization hyperparameters and learning-rate schedule.

\begin{table}[t]
\centering
\caption{Optimization hyperparameters used for deterministic-backbone
and pseudo-ensemble-head training.}
\label{tab:supp_hparams}
\small
\begin{tabular}{lc}
\toprule
\textbf{Hyperparameter} & \textbf{Value} \\
\midrule
Optimizer & SGD \\
Momentum & $0.9$ \\
Initial learning rate & $0.1$ \\
Weight decay & $10^{-4}$ \\
Learning-rate schedule & Cosine annealing \\
Minimum learning rate & $0$ \\
Batch size & $64$ \\
Input resolution & $64\times64$ \\
Backbone epochs & $300$ \\
Head-only epochs & $100$ \\
Simulation steps & $T=2$ \\
\bottomrule
\end{tabular}
\end{table}

\subsection{Baseline Implementations}
\label{sec:supp_baselines}

We compare against deterministic confidence and feature-distance
methods, stochastic-head approximations, and conventional deep
ensembles.

\paragraph{Deterministic MSP and entropy.}
MSP and predictive entropy are computed directly from the softmax
distribution of a deterministically trained SNN.

\paragraph{Deep ensemble.}\cite{lakshminarayanan2017ensembles}
Deep-ensemble predictions are obtained by averaging the class
probabilities of independently initialized and trained deterministic
SNNs. Each ensemble member contains a complete backbone and
classification head.

\paragraph{Monte Carlo dropout.}\cite{gal2016dropout}
For MC dropout, we attach and retrain a classification head containing
dropout with probability $p=0.2$ after its LIF activation. The head is
trained for 100 epochs, and uncertainty is estimated using
$T_{\mathrm{MC}}=20$ stochastic inference passes.

\paragraph{Last-layer Laplace} \cite{daxberger2021laplace}
We fit a Kronecker-factored Laplace approximation to the final
classification layer. For Spikformer, the posterior is fitted to
\texttt{backbone.head}, corresponding to
$\mathrm{Linear}(384,C)$. For ResNet19-SNN, it is fitted to
\texttt{model.fc2}, corresponding to $\mathrm{Linear}(256,C)$.
The prior precision is optimized using marginal likelihood, and
predictions are averaged over $S=20$ posterior samples.

\paragraph{Mahalanobis distance~\cite{lee2018mahalanobis}.}
Mahalanobis detection uses temporal-mean-pooled firing-rate features
from the penultimate representation. These features are
384-dimensional for Spikformer and 512-dimensional for ResNet19-SNN.
We estimate one mean per ID class and a covariance matrix tied across
classes. The OOD score is the distance to the nearest class mean.


\subsection{Evaluation Protocol}
\label{sec:supp_evaluation_protocol}

We train five independently initialized deterministic backbones using
the fixed random seeds
\[
\{0,1,2,3,4\}.
\]
For a configuration containing $K_b$ independently trained backbones,
we evaluate every subset among the five available models. Results for
$K_b=1$, $K_b=2$, and $K_b=3$ are therefore averaged over 5, 10, and
10 backbone subsets, respectively. For $K_b=5$, only one subset exists
and its reported subset standard deviation is zero. Each selected
pseudo-ensemble backbone contributes all of its $K_h=5$ heads, whereas
a deep-ensemble member has one head.

Post-hoc single-backbone baselines are summarized across the five
deterministic-backbone seeds. The hyperparameter ablation in
Section \textit{Hyperparameter Ablation} uses three deterministic backbones,
corresponding to seeds $\{0,1,2\}$.

We report EuroSAT test accuracy and OOD detection using MSP,
predictive entropy,  and mutual information.
For AUROC and AUPR, OOD samples are treated as the positive class.
All reported metric values are percentages unless stated otherwise.
Higher accuracy, AUROC, and AUPR are better, whereas lower FPR@95 is
better.

\subsection{Uncertainty Scores and OOD Metrics}
\label{sec:uncertainty_metrics}

\paragraph{Predictive distribution.}
Let $J$ denote the total number of predictive members and $C$ the
number of classes. Member $j$ produces
$\mathbf p^{(j)}(x)\in\mathbb R^C$, and the aggregated distribution is
\begin{equation}
\bar p_c(x)
=
\frac{1}{J}\sum_{j=1}^{J}p_c^{(j)}(x),
\qquad c=1,\ldots,C.
\label{eq:supp_mean_predictive_distribution}
\end{equation}
For a pseudo-ensemble with $K_b$ backbones and $K_h$ heads per
backbone, $J=K_bK_h$. For a conventional deep ensemble, $J=K_b$.
For MC dropout and last-layer Laplace, $J$ denotes the number of
stochastic predictive samples.

\paragraph{MSP-based uncertainty.}
We orient the MSP score so that larger values indicate stronger OOD
evidence:
\begin{equation}
s_{\mathrm{MSP}}(x)
=
1-\max_c\bar p_c(x).
\label{eq:supp_msp_score}
\end{equation}

\paragraph{Predictive entropy.}
\begin{equation}
s_{\mathrm{Ent}}(x)
=
H\!\left(\bar{\mathbf p}(x)\right)
=
-\sum_{c=1}^{C}\bar p_c(x)\log\bar p_c(x).
\label{eq:supp_predictive_entropy}
\end{equation}


\paragraph{Mutual information.}
\begin{equation}
s_{\mathrm{MI}}(x)
=
H\!\left(\bar{\mathbf p}(x)\right)
-
\frac{1}{J}\sum_{j=1}^{J}H\!\left(\mathbf p^{(j)}(x)\right).
\label{eq:supp_mutual_information}
\end{equation}
Mutual information is high when individual predictors are confident
but mutually inconsistent.

\paragraph{Mahalanobis distance.}
Let $\mathbf z(x)$ denote the temporal-mean-pooled penultimate feature,
$\boldsymbol\mu_c$ the feature mean of class $c$, and
$\boldsymbol\Sigma$ the covariance matrix shared across classes. The
score is
\begin{equation}
s_{\mathrm{Maha}}(x)
=
\min_c
\left(\mathbf z(x)-\boldsymbol\mu_c\right)^\top
\boldsymbol\Sigma^{-1}
\left(\mathbf z(x)-\boldsymbol\mu_c\right).
\label{eq:supp_mahalanobis}
\end{equation}

\begin{table*}[t]
\centering
\caption{
Complete diagnostic comparison of image transformations on EuroSAT
using a conventional deep ensemble of five independently trained
Spikformer models. Values denote the mean $\pm$ standard deviation
across evaluation samples. Lower MSP indicates lower predictive
confidence, whereas higher predictive entropy, predictive variance,
and mutual information indicate greater uncertainty or inter-model
disagreement. Bold and underlined values denote the strongest and
second-strongest uncertainty responses, respectively.
}

\label{tab:supp_corruption_diagnostic}
\small
\setlength{\tabcolsep}{3.0pt}
\renewcommand{\arraystretch}{1.12}

\resizebox{\textwidth}{!}{%
\begin{tabular}{@{}lcccccccccc@{}}
\toprule
Metric
& Clean
& Noise
& Blur-S
& Blur-M
& Blur-L
& Bright.
& Dark
& Contrast
& Rot. $90^\circ$
& Cutout \\
\midrule

MSP $\downarrow$
& \ms{0.974}{0.080}
& \ms{0.760}{0.209}
& \ms{0.779}{0.190}
& \second{0.712}{0.188}
& \best{0.687}{0.177}
& \ms{0.769}{0.201}
& \ms{0.839}{0.190}
& \ms{0.785}{0.210}
& \ms{0.965}{0.093}
& \ms{0.941}{0.121}
\\

Predictive entropy $\uparrow$
& \ms{0.076}{0.183}
& \ms{0.650}{0.480}
& \ms{0.557}{0.401}
& \second{0.707}{0.365}
& \best{0.752}{0.334}
& \ms{0.611}{0.455}
& \ms{0.473}{0.441}
& \ms{0.602}{0.499}
& \ms{0.104}{0.219}
& \ms{0.165}{0.271}
\\

Predictive variance $\uparrow$
& \ms{0.0012}{0.0048}
& \ms{0.0131}{0.0132}
& \ms{0.0113}{0.0114}
& \second{0.0154}{0.0118}
& \best{0.0172}{0.0117}
& \ms{0.0128}{0.0126}
& \ms{0.0085}{0.0119}
& \ms{0.0106}{0.0119}
& \ms{0.0017}{0.0057}
& \ms{0.0029}{0.0072}
\\

Mutual information $\uparrow$
& \ms{0.0234}{0.0738}
& \ms{0.2522}{0.2223}
& \ms{0.1968}{0.1762}
& \second{0.2627}{0.1779}
& \best{0.2877}{0.1732}
& \ms{0.2382}{0.2102}
& \ms{0.1745}{0.1990}
& \ms{0.2128}{0.2089}
& \ms{0.0329}{0.0893}
& \ms{0.0531}{0.1105}
\\

\bottomrule
\end{tabular}%
}
\end{table*}

\begin{table*}[t]
\centering
\caption{
OOD detection using predictive entropy with EuroSAT as ID.
ADPE denotes our proposed method and is identified by a bold method
name. AUPR treats OOD as the positive class.
All values are percentages.
Higher AUROC and AUPR are better; lower FPR@95 is better.
The best and second-best results within each backbone and metric
column are shown in bold and underlined, respectively.
}
\label{tab:ucm_aid_entropy}
\small
\setlength{\tabcolsep}{4.2pt}
\renewcommand{\arraystretch}{1.07}

\begin{tabular}{@{}lcccccc@{}}
\toprule
\multirow{2}{*}{Method}
& \multicolumn{3}{c}{UCM}
& \multicolumn{3}{c}{AID}
\\
\cmidrule(lr){2-4}
\cmidrule(lr){5-7}
& AUROC$\uparrow$
& AUPR$\uparrow$
& FPR@95$\downarrow$
& AUROC$\uparrow$
& AUPR$\uparrow$
& FPR@95$\downarrow$
\\
\midrule

\multicolumn{7}{l}{\textbf{Spikformer}}\\
\midrule

MC-DO $(1,\text{--})$
& \ms{92.31}{0.49}
& \ms{91.30}{0.64}
& \ms{33.75}{2.94}
& \ms{93.87}{0.99}
& \ms{98.17}{0.29}
& \ms{29.74}{3.95}
\\

LLL $(1,\text{--})$
& \ms{92.22}{1.46}
& \ms{91.22}{1.60}
& \ms{34.47}{4.00}
& \ms{93.23}{1.36}
& \ms{97.94}{0.47}
& \ms{32.71}{6.00}
\\
\midrule

DE $(1,1)$
& \ms{89.99}{1.76}
& \ms{88.75}{1.78}
& \ms{41.41}{3.58}
& \ms{91.04}{2.12}
& \ms{97.21}{0.72}
& \ms{40.29}{6.91}
\\

DE $(5,1)$
& \ms{95.51}{0.00}
& \ms{95.14}{0.00}
& \ms{21.29}{0.00}
& \ms{96.19}{0.00}
& \ms{98.89}{0.00}
& \ms{19.28}{0.00}
\\
\midrule

CEPE $(3,5)$
& \ms{95.60}{0.22}
& \ms{95.42}{0.24}
& \ms{20.16}{0.90}
& \ms{96.55}{0.31}
& \ms{99.03}{0.10}
& \ms{17.49}{1.72}
\\
\midrule

\textbf{ADPE $(1,5)$}
& \ms{94.92}{0.60}
& \ms{94.30}{0.77}
& \ms{22.72}{2.49}
& \ms{95.40}{1.06}
& \ms{98.62}{0.35}
& \ms{21.85}{5.40}
\\

\textbf{ADPE $(2,5)$}
& \second{96.86}{0.41}
& \second{96.53}{0.41}
& \second{15.61}{1.43}
& \second{97.15}{0.45}
& \second{99.19}{0.15}
& \second{14.04}{2.39}
\\

\textbf{ADPE $(3,5)$}
& \best{97.51}{0.26}
& \best{97.25}{0.23}
& \best{12.59}{0.71}
& \best{97.68}{0.23}
& \best{99.36}{0.08}
& \best{11.21}{1.17}
\\

\midrule
\multicolumn{7}{l}{\textbf{ResNet19-SNN}}\\
\midrule

MC-DO $(1,\text{--})$
& \ms{89.06}{1.20}
& \ms{87.83}{1.42}
& \ms{41.35}{5.40}
& \ms{90.53}{0.88}
& \ms{97.09}{0.32}
& \ms{38.72}{4.63}
\\

LLL $(1,\text{--})$
& \ms{89.88}{0.46}
& \ms{88.57}{0.78}
& \ms{41.06}{2.83}
& \ms{90.82}{0.54}
& \ms{97.19}{0.19}
& \ms{39.03}{3.31}
\\
\midrule

DE $(1,1)$
& \ms{89.22}{0.97}
& \ms{87.49}{1.25}
& \ms{43.27}{3.43}
& \ms{90.24}{0.57}
& \ms{96.92}{0.21}
& \ms{41.37}{2.96}
\\

DE $(5,1)$
& \second{92.28}{0.00}
& \second{91.64}{0.00}
& \second{31.71}{0.00}
& \second{93.22}{0.00}
& \second{98.00}{0.00}
& \second{30.10}{0.00}
\\
\midrule

CEPE $(3,5)$
& \ms{91.69}{0.18}
& \ms{91.13}{0.26}
& \ms{32.88}{1.10}
& \ms{93.00}{0.15}
& \ms{97.95}{0.06}
& \ms{30.38}{1.15}
\\
\midrule

\textbf{ADPE $(1,5)$}
& \ms{90.14}{0.74}
& \ms{89.31}{1.06}
& \ms{37.45}{3.51}
& \ms{91.30}{0.53}
& \ms{97.36}{0.19}
& \ms{36.82}{2.33}
\\

\textbf{ADPE $(2,5)$}
& \ms{91.86}{0.33}
& \ms{91.32}{0.51}
& \ms{32.60}{1.88}
& \ms{92.92}{0.29}
& \ms{97.90}{0.11}
& \ms{31.48}{1.45}
\\

\textbf{ADPE $(3,5)$}
& \best{92.48}{0.17}
& \best{92.04}{0.31}
& \best{30.40}{1.33}
& \best{93.51}{0.19}
& \best{98.09}{0.07}
& \best{29.24}{1.46}
\\

\bottomrule
\end{tabular}
\end{table*}



\begin{table*}[t]
\centering
\caption{
Mutual-information-based AUPR-Out for ensemble-based methods with
EuroSAT as ID.
The pair $(K_b,K_h)$ denotes the number of independently trained
backbones and the number of heads per backbone.
Results are reported as mean$\pm$standard deviation over the evaluated
backbone combinations. All values are percentages, and higher values
are better. The best and second-best results within each backbone and
dataset are shown in bold and underlined, respectively.
}
\label{tab:mi_aupr}
\small
\setlength{\tabcolsep}{7pt}
\renewcommand{\arraystretch}{1.08}

\begin{tabular}{@{}lcccc@{}}
\toprule
Method
& UCM
& AID
& Global-Near
& Global-Far \\
\midrule

\multicolumn{5}{l}{\textbf{Spikformer}}\\
\midrule

DE $(5,1)$
& \best{94.30}{0.00}
& \ms{98.62}{0.00}
& \ms{96.51}{0.00}
& \second{94.17}{0.00}
\\

CEPE $(3,5)$
& \ms{89.41}{0.94}
& \ms{97.75}{0.24}
& \ms{96.16}{0.40}
& \ms{92.12}{0.40}
\\

\textbf{ADPE $(1,5)$}
& \ms{93.52}{0.66}
& \ms{98.48}{0.32}
& \ms{95.70}{0.67}
& \ms{92.00}{1.07}
\\

\textbf{ADPE $(2,5)$}
& \ms{93.37}{1.03}
& \second{98.72}{0.19}
& \second{96.77}{0.42}
& \ms{93.74}{1.14}
\\

\textbf{ADPE $(3,5)$}
& \second{94.05}{0.78}
& \best{98.90}{0.11}
& \best{97.24}{0.22}
& \best{94.72}{0.81}
\\

\midrule
\multicolumn{5}{l}{\textbf{ResNet19-SNN}}\\
\midrule

DE $(5,1)$
& \ms{90.17}{0.00}
& \best{97.68}{0.00}
& \best{96.58}{0.00}
& \best{95.91}{0.00}
\\

CEPE $(3,5)$
& \ms{88.98}{0.40}
& \ms{97.46}{0.10}
& \second{96.24}{0.25}
& \ms{95.13}{0.67}
\\

\textbf{ADPE $(1,5)$}
& \ms{88.70}{1.17}
& \ms{96.73}{0.34}
& \ms{94.85}{0.61}
& \ms{93.14}{0.89}
\\

\textbf{ADPE $(2,5)$}
& \second{90.20}{0.40}
& \ms{97.32}{0.14}
& \ms{95.67}{0.47}
& \ms{94.96}{0.93}
\\

\textbf{ADPE $(3,5)$}
& \best{91.03}{0.18}
& \second{97.59}{0.09}
& \ms{96.01}{0.34}
& \second{95.78}{0.62}
\\

\bottomrule
\end{tabular}
\end{table*}
\paragraph{AUROC, AUPR-Out, and FPR@95.}
All scores are oriented so that larger values indicate stronger OOD
evidence. AUROC measures separation across all thresholds, and
AUPR-Out is the area under the precision--recall curve with OOD as the
positive class. For FPR@95, we select the score threshold
\begin{equation}
\tau_{95}
=
Q_{0.95}
\left(
\{s(x):x\in\mathcal D_{\mathrm{ID}}\}
\right),
\label{eq:supp_fpr_threshold}
\end{equation}
which accepts 95\% of the ID samples. Because larger scores indicate
OOD evidence, an example is accepted as ID when
$s(x)\leq\tau_{95}$. We compute
\begin{equation}
\mathrm{FPR@95}
=
\frac{1}{|\mathcal D_{\mathrm{OOD}}|}
\sum_{x\in\mathcal D_{\mathrm{OOD}}}
\mathbb I\!\left[s(x)\leq\tau_{95}\right].
\label{eq:supp_fpr95}
\end{equation}

\subsection{Implementation and Reproducibility}
\label{sec:supp_implementation}

All experiments are implemented in PyTorch 2.6.0 using SpikingJelly
0.0.0.0.14 \citep{fang2023spikingjelly} and are executed on NVIDIA A100 GPUs. Each seed controls
the Python, NumPy, PyTorch CPU, and PyTorch CUDA random states. CuDNN
deterministic execution is enabled with
\texttt{deterministic=True} and \texttt{benchmark=False}. The same
dataset split is reused for all methods.

\subsection{Parameter and Computation Accounting}
\label{sec:supp_cost}

For a configuration with $K_b$ backbones and $K_h$ heads per
backbone, the parameter count and inference-computation proxy are
\begin{equation}
P_{\mathrm{ADPE}}
=
K_b(P_b+K_hP_h),
\qquad
\mathcal C_{\mathrm{ADPE}}
=
K_b(\mathcal C_b+K_h\mathcal C_h),
\end{equation}
where $P_b$ and $\mathcal C_b$ denote the parameters and computation
of one backbone, and $P_h$ and $\mathcal C_h$ denote those of one
head. For both architectures, $P_b\approx12.5$ million and
$P_h\approx0.1$ million, so five heads add approximately 4\% parameter
overhead to one backbone. Because the head computation is small
relative to the backbone computation, the dominant deployment cost
scales with $K_b$ rather than $K_bK_h$. We therefore report parameter
storage and the number of backbone evaluations as deployment-cost
proxies.

\section{Diagnostic Study of Structured Disagreement Inputs}
\label{sec:supp_corruption_study}

Before training ADPE, we evaluate a conventional deep ensemble of five
independently trained Spikformer models on transformed EuroSAT images.
This diagnostic study is used only to identify structured inputs that
naturally induce inter-model uncertainty and disagreement. The
transformed images are not treated as samples from the unknown
test-time OOD distributions.

\subsection{Evaluated Transformations}

We consider additive noise (gaussian noise with std =0.3), small(window size=5), medium(9), and large box blur(11),
brightness increase(0.5), darkening(-0.5), contrast (factor=2) modification, a
$90^\circ$ rotation, and cutout(16x16). The three blur levels correspond to
increasing box-filter kernel sizes which use average-pooling kernels. Medium and large blur  produce the
strongest predictive-entropy and mutual-information responses while
retaining coarse scene structure, motivating their use as structured
disagreement inputs.
Table~\ref{tab:supp_corruption_diagnostic} provides the complete
comparison underlying the transformation-selection study in the main
paper. Clean images produce high confidence and little inter-model
disagreement. Rotation and cutout cause only small changes in all four
diagnostics, indicating that the ensemble remains comparatively
confident under these transformations. Additive noise, brightness,
darkening, and contrast modification induce greater uncertainty, but
medium and large box blur produce the most consistent response across
all measures. In particular, large blur yields the lowest MSP and the
highest predictive entropy, predictive variance, and mutual
information. We therefore use box-blurred images as structured
disagreement inputs for ADPE.
\paragraph{Predictive variance.}
The class-wise variance across predictive members is averaged over
classes:
\begin{equation}
s_{\mathrm{Var}}(x)
=
\frac{1}{C}
\sum_{c=1}^{C}
\left[
\frac{1}{J}
\sum_{j=1}^{J}
\left(p_c^{(j)}(x)-\bar p_c(x)\right)^2
\right].
\label{eq:supp_predictive_variance}
\end{equation}
\section{Additional OOD Detection Results}
\label{sec:supp_additional_results}



Tables~\ref{tab:ucm_aid_entropy} and~\ref{tab:mi_aupr} provide
complementary evidence that the proposed disagreement objective
improves both predictive uncertainty and inter-member diversity.
In Table~\ref{tab:ucm_aid_entropy}, ADPE $(3,5)$ consistently
outperforms CEPE $(3,5)$ for predictive-entropy-based OOD detection
on UCM and AID across both backbones. The improvement is especially
pronounced for Spikformer, where FPR@95 decreases from $20.16\%$ to
$12.59\%$ on UCM and from $17.49\%$ to $11.21\%$ on AID. For
ResNet19-SNN, the gains are smaller but remain consistent across
AUROC, AUPR, and FPR@95. Table~\ref{tab:mi_aupr} further shows that
these improvements are accompanied by more informative predictive
disagreement. Spikformer ADPE $(3,5)$ achieves the highest
MI-based AUPR-Out on AID, Global-Near, and Global-Far, while remaining
close to the five-model deep ensemble on UCM. For ResNet19-SNN,
ADPE $(3,5)$ obtains the strongest UCM result and remains competitive
with the deep ensemble on the remaining distribution shifts. Together,
the two tables indicate that ADPE improves uncertainty quality not only
by modifying the ensemble-averaged confidence, but also by increasing
useful diversity among the predictive members.


\section{Hyperparameter Ablation}
\label{sec:supp_ablation}

\subsection{Effect of the Number of Heads and Disagreement Weight}

We vary the number of heads
$K_h\in\{2,3,4,5,6,7\}$ and the disagreement weight
$\lambda_{\mathrm{dis}}\in\{0.1,0.2,0.3,0.4,0.5\}$ for
ResNet19-SNN. Every entry is the mean and sample standard deviation
over deterministic-backbone seeds $\{0,1,2\}$. The setting used in the
main experiments, $K_h=5$ and $\lambda_{\mathrm{dis}}=0.3$, is shown in
bold, and the best mean is underlined.

\begin{table*}[t]
\centering
\caption{Ablation of the number of heads $K_h$ and disagreement weight
$\lambda_{\mathrm{dis}}$ on UCM using MSP-based FPR@95. Results are
mean $\pm$ standard deviation over three independently trained
backbones. Lower values are better.}
\label{tab:ablation_msp_fpr95}
\small
\setlength{\tabcolsep}{6pt}
\renewcommand{\arraystretch}{1.15}

\begin{tabular}{c*{6}{c}}
\toprule
$\lambda_{\mathrm{dis}}\backslash K_h$
& 2 & 3 & 4 & 5 & 6 & 7 \\
\midrule
0.1
& $39.33\pm4.33$
& $39.60\pm3.01$
& $38.51\pm4.61$
& $37.86\pm3.65$
& $37.22\pm4.84$
& $37.98\pm4.83$ \\

0.2
& $40.41\pm3.96$
& $38.67\pm4.77$
& $38.27\pm3.27$
& $37.63\pm4.36$
& $37.40\pm4.42$
& $37.79\pm3.76$ \\

0.3
& $40.52\pm5.36$
& $38.84\pm3.74$
& $37.97\pm4.48$
& $\mathbf{38.00\pm4.49}$
& $38.25\pm4.91$
& $39.19\pm4.84$ \\

0.4
& $40.10\pm5.03$
& $38.78\pm3.59$
& $37.09\pm4.68$
& $37.57\pm4.67$
& $37.71\pm4.87$
& $37.06\pm6.01$ \\

0.5
& $39.22\pm4.85$
& $38.79\pm4.28$
& $38.83\pm4.73$
& $\underline{36.92\pm5.01}$
& $37.54\pm4.08$
& $37.62\pm5.17$ \\
\bottomrule
\end{tabular}

\vspace{1mm}
{\footnotesize Bold indicates the configuration used in the main
experiments; underlining indicates the best mean result.}
\end{table*}

\begin{table*}[t]
\centering
\caption{Ablation of the number of heads $K_h$ and disagreement weight
$\lambda_{\mathrm{dis}}$ on UCM using predictive-entropy-based
FPR@95. Results are mean $\pm$ standard deviation over three
independently trained backbones. Lower values are better.}
\label{tab:ablation_entropy_fpr95}
\small
\setlength{\tabcolsep}{6pt}
\renewcommand{\arraystretch}{1.15}

\begin{tabular}{c*{6}{c}}
\toprule
$\lambda_{\mathrm{dis}}\backslash K_h$
& 2 & 3 & 4 & 5 & 6 & 7 \\
\midrule
0.1
& $37.94\pm4.66$
& $37.29\pm2.97$
& $36.73\pm3.78$
& $35.97\pm3.24$
& $35.70\pm4.23$
& $35.35\pm3.48$ \\

0.2
& $38.84\pm3.78$
& $36.79\pm4.77$
& $35.68\pm2.54$
& $36.00\pm4.18$
& $35.22\pm3.11$
& $35.52\pm3.27$ \\

0.3
& $39.03\pm4.68$
& $36.98\pm3.41$
& $35.30\pm3.75$
& $\mathbf{35.81\pm4.07}$
& $35.90\pm3.52$
& $36.22\pm3.62$ \\

0.4
& $38.47\pm5.62$
& $36.78\pm4.06$
& $35.44\pm4.39$
& $35.37\pm4.72$
& $35.53\pm4.47$
& $\underline{34.65\pm5.02}$ \\

0.5
& $37.70\pm4.76$
& $36.33\pm3.87$
& $36.09\pm3.96$
& $35.10\pm4.75$
& $35.62\pm3.98$
& $35.03\pm4.28$ \\
\bottomrule
\end{tabular}

\vspace{1mm}
{\footnotesize Bold indicates the configuration used in the main
experiments; underlining indicates the best mean result.}
\end{table*}

The ablation exhibits a broad region of competitive performance rather
than a sharply localized optimum. The selected configuration,
$K_h=5$ and $\lambda_{\mathrm{dis}}=0.3$, is within approximately
$1.2$ percentage points of the best mean FPR@95 for both MSP and
predictive entropy. These differences are smaller than the observed
cross-seed variability, indicating that the selected configuration is
a robust intermediate setting without requiring the largest number of
heads or the strongest disagreement weight.






\end{document}